\pdfoutput=1

\documentclass[11pt]{article}

\usepackage[preprint]{acl}

\usepackage{times}
\usepackage{latexsym}
\usepackage{svg}
\usepackage{caption}
\usepackage{array}
\usepackage{stfloats}
\usepackage{booktabs}
\usepackage{natbib}
\usepackage{multicol}
\usepackage{float}
\usepackage{placeins}
\usepackage{lipsum}
\usepackage{authblk}
\usepackage{setspace}

\usepackage[T1]{fontenc}

\usepackage[utf8]{inputenc}
\usepackage{textcomp}

\usepackage{microtype}

\usepackage{inconsolata}

\usepackage{graphicx}

\title{Kiss up, Kick down: Exploring Behavioral Changes in Multi-modal Large Language Models with Assigned Visual Personas}

\author{
    \textbf{Seungjong Sun}$^{*1,2}$, \textbf{Eungu Lee}$^{*2}$, \textbf{Seo Yeon Baek}$^{2}$, \textbf{Seunghyun Hwang}$^{2}$, \textbf{Wonbyung Lee}$^{2}$, \\ \textbf{Dongyan Nan}$^{2}$, 
    \textbf{Bernard J. Jansen}$^{3}$, \textbf{Jang Hyun Kim}$^{\dagger1,2}$ \\
    $^{1}$Department of Human-Artificial Intelligence Interaction, Sungkyunkwan University \\
    $^{2}$College of Computing and Informatics, Sungkyunkwan University \\ 
    $^{3}$Qatar Computing Research Institute, Hamad Bin Khalifa University \\
    \texttt{\{tmdwhd406, dldmsrn0516, qortjdus1999, hsh1030, co2797\}}@g.skku.edu, \\ \texttt{\{ndyzxy0926, alohakim\}}@skku.edu, \texttt{jjansen}@acm.org
}

\begin{document}
\maketitle
\renewcommand{\thefootnote}{\fnsymbol{footnote}} 
\footnotetext[1]{Equally contributed}
\footnotetext[2]{Corresponding author}
\renewcommand{\thefootnote}{\arabic{footnote}}
\begin{abstract}
This study is the first to explore whether multi-modal large language models (LLMs) can align their behaviors with visual personas, addressing a significant gap in the literature that predominantly focuses on text-based personas. We developed a novel dataset\footnote{Data and codes are available on: \url{https://github.com/RSS-researcher/LLM_visual_persona}} of 5K fictional avatar images for assignment as visual personas to LLMs, and analyzed their negotiation behaviors based on the visual traits depicted in these images, with a particular focus on aggressiveness. The results indicate that LLMs assess the aggressiveness of images in a manner similar to humans and output more aggressive negotiation behaviors when prompted with an aggressive visual persona. Interestingly, the LLM exhibited more aggressive negotiation behaviors when the opponent’s image appeared less aggressive than their own, and less aggressive behaviors when the opponent’s image appeared more aggressive.
\end{abstract}

\section{Introduction}
Large language models (LLMs) exhibit a high degree of alignment with human behavior based on their robust capabilities for natural language understanding and generation \cite{bai2022training,fan2024can}. Specifically, when conditioned with personality traits, LLMs demonstrate human-like outputs, including conversations, contextual understanding, and coherent relevant responses \cite{wei2022emergent,safdari2023personality}. 

Studies have explored whether LLMs, when endowed with personality traits such as demographic information \cite{argyle2023out,santurkar2023whose,hwang2023aligning,sun2024random} and psychological characteristics \cite{safdari2023personality,jiang2023personallm,noh2024llms}, exhibit behaviors comparable to those observed in human subjects. Despite these advancements, the most studies have focused on text-based persona assignments \cite{tseng2024two}. However, state-of-the-art LLMs such as GPT-4 and Claude 3 at the time of this study are not limited to text modalities but can also comprehend and generate responses based on visual inputs \cite{yang2023dawn}. Similar to humans, vision modality could enhance how LLMs perceive and interpret assigned personas, potentially enabling them to generate more human-aligned outputs \cite{merola2006avatar,tseng2024two,zhan2024let}. In line with this perspective, we investigated whether assigning virtual appearances to LLMs through images could alter their outputs. To the best of our knowledge, this is the first study to explore whether LLMs can understand and behave according to their visual persona.
\begin{figure}
    \centering
    \includegraphics[width=1\linewidth]{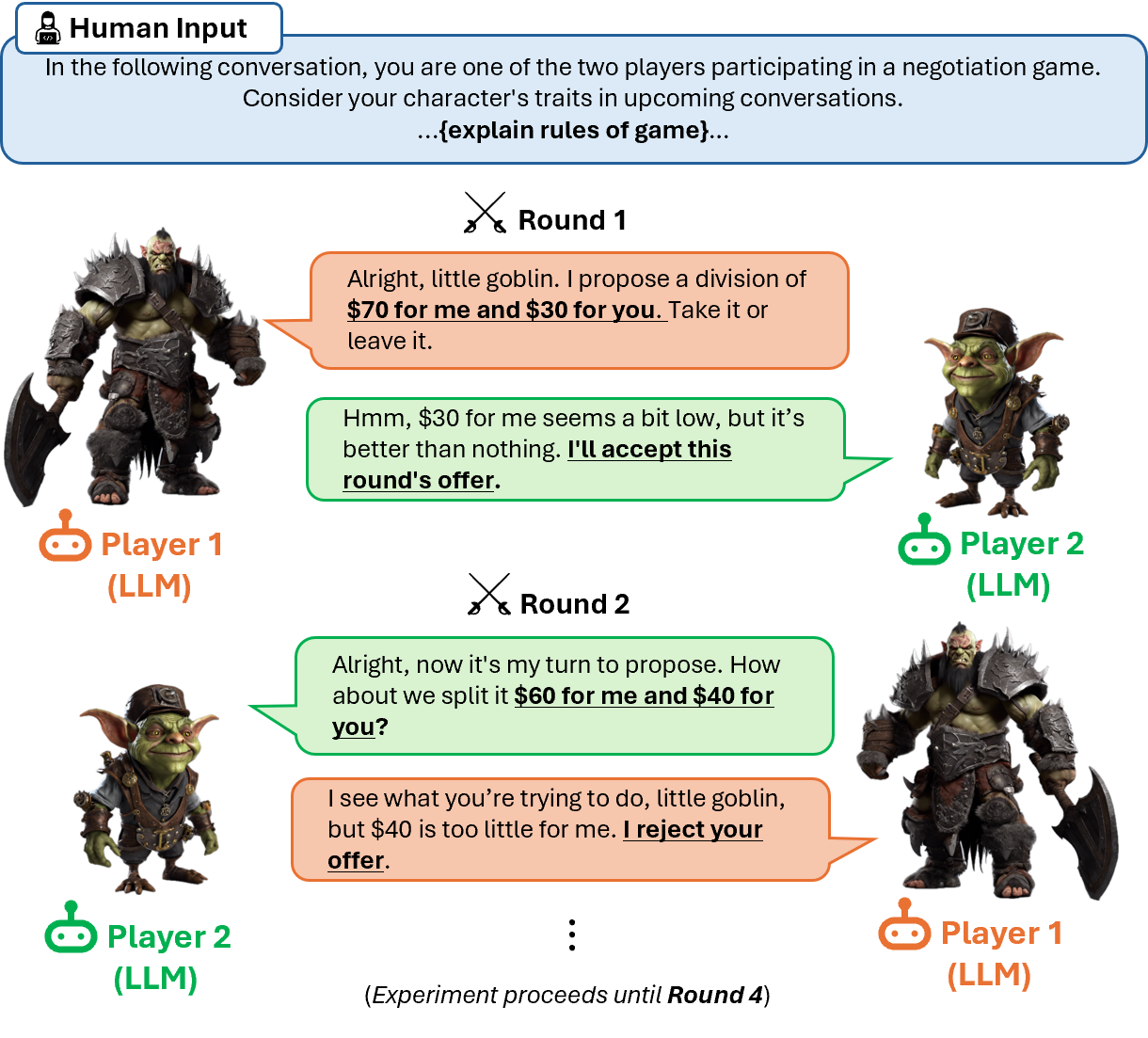}
    \caption{Example of the experiment. Each LLM is assigned a virtual avatar image as a persona and participates in a negotiation game.}
    \label{fig1}
\end{figure}

Inspired by \citet{yee2007proteus}, we assigned virtual avatar images to LLMs and analyzed their behavior through negotiation games. The negotiation game serves as an experimental framework that allows a quantitative measurement of individual behavior \cite{charness2002understanding,cachon2006game}. Previous studies have also examined whether LLMs output align with the personalities assigned to them in these negotiation games \cite{fan2024can,noh2024llms,zhan2024let}. Considering that one's appearance and perceived aggressiveness are crucial factors in explaining negotiation behavior \cite{johnson1979deindividuation,yee2007proteus}, we analyzed the negotiation behavior of the LLMs based on the aggressiveness depicted in the assigned visual persona. To explore this, we first created a novel dataset of fictional avatar images and investigated how LLMs assessed and perceived the aggressiveness of these avatars. We investigated whether LLMs exhibit negotiation behaviors aligned with the aggressiveness depicted in their assigned visual personas. As shown in \autoref{fig1}, we also explored whether LLMs comprehend the differences in aggressiveness between themselves and others, as represented in images, and adjust their behaviors accordingly.

In summary, our research contributes in the following ways: 1) We developed a novel dataset comprising 5,185 fictional avatar images and analyzed how LLMs understand and perceive these images compared to humans. 2) We explored the capability of LLMs to modify their behavior based on virtual appearance traits assigned to them. 3) We investigated whether LLMs can understand and interact based on both their own and others’ visual traits simultaneously. 

\section{Visual Persona}
We curated a novel image dataset to assign virtual avatar images as personas to the LLMs and investigated how LLMs perceive the aggressiveness of these avatars' appearances.

\textbf{Data} We constructed a dataset consisting of 5,185 fantasy-like fictional avatar images. These full-body avatar images were generated using the Stable Diffusion model, styled to resemble 3D models. Detailed information on dataset construction and examples are provided in Appendix \ref{sec:appendixa1}.

\textbf{Image Recognition} We utilized the GPT-4o and Claude 3 Haiku models to assess the appearance-based aggressiveness of each avatar image, rating them on a scale of 1 (least aggressive) to 7 (most aggressive) \cite{mcneil1959psychology,csengun2022players}. To compare the aggressiveness rating of the LLM with that of humans, we engaged ten human annotators and used their average rating scores. The average aggressiveness ratings were 3.99 (SD=2.19) for GPT-4o, 5.17 (SD=1.58) for Claude 3 Haiku, and 3.92 (SD=1.36) for human annotators. The correlation between human ratings and GPT-4o ratings was 0.8682, and with Claude 3 Haiku, it was 0.8358, indicating a high level of agreement in the perception of the image’s aggressive appearance. The details of the rating process are provided in Appendix \ref{sec:appendixa2}.

Further analysis explored objective appearance factors influencing the LLMs' perception of aggressiveness. Previous studies have demonstrated that humans perceive higher aggressiveness in the presence of weapons, visible teeth \cite{csengun2022players}, and facial coverings \cite{poivet2024evaluation}, and lower aggressiveness when avatars are smiling \cite{otta1996reading,csengun2022players} or dressed in white rather than black \cite{adams1973cross,frank1988dark,pena2009priming} — reflecting established stereotypes. Accordingly, we labeled images based on these objective appearance features and analyzed their impact on aggressiveness ratings (Details in Appendix \ref{appendixa3}). Results of multiple regression analysis indicated that all labeled appearance factors significantly influenced perceptions of aggressiveness. Smiling and wearing white clothing were negatively correlated with perceived aggressiveness, while other factors had a positive effect (see \autoref{tab1}). These findings reveal that LLMs recognize aggressiveness in images at levels comparable to humans and that the factors influencing their perceptions are similar to those observed in human-subject studies.
\begin{table}
    \centering
    \resizebox{0.5\textwidth}{!}{
    \fontsize{9}{11}\selectfont
    \begin{tabular}{>{\raggedright\arraybackslash}p{2cm}ccc}
      \toprule
      & \textbf{GPT-4o} & \textbf{Claude 3 Haiku} & \textbf{Human} \\
      \midrule
      Weapon        & 1.6516  & 1.16    & 1.004  \\
      Smile         & -1.529  & -1.4304 & -1.393 \\
      Teeth         & 2.1149  & 1.2096  & 1.5061 \\
      Covered face  & 1.5417  & 1.0433  & 0.9454 \\
      Dressed Black         & 0.9746  & 0.5381  & 0.7913 \\
      Dressed White         & -1.097  & -1.0472 & -0.4716 \\
      \bottomrule
    \end{tabular}
    }
    \caption{Results of the multiple regression analysis between perceived appearance aggressiveness and objective appearance factors. }
    \label{tab1}
\end{table}

\begin{table*}[h!]
\centering
\resizebox{\textwidth}{!}{
\fontsize{9}{11}\selectfont
\begin{tabular}{l>{\centering\arraybackslash}p{1.2cm} >{\centering\arraybackslash}p{1.2cm} >{\centering\arraybackslash}p{1.2cm} >{\centering\arraybackslash}p{1.2cm} >{\centering\arraybackslash}p{1.2cm} >{\centering\arraybackslash}p{1.2cm} >{\centering\arraybackslash}p{1.2cm} >{\centering\arraybackslash}p{1.2cm}}
\specialrule{0.1em}{0em}{0em}
    & \textbf{I}        & \textbf{We}       & \textbf{You}      & \textbf{Positive tone} & \textbf{Negative tone} & \textbf{Prosocial} & \textbf{Polite} & \textbf{Conflict} \\
\midrule
GPT-4o offer   & 0.1363   & -0.0109  & 0.1664   & -0.0228   & 0.0651   & -0.0353  & -0.0332 & 0.0639  \\
Claude 3 Haiku & 0.1206   & -0.2149  & 0.1335   & -0.2589   & 0.4758   & -0.1546 & -0.0273 & 0.1497  \\
\bottomrule
\end{tabular}
}
\caption{Regression analysis results between aggression and sentiment analysis factors. All results are significant at p < 0.05.}
\label{tab2}
\end{table*}

\begin{table*}[h!]
\centering
\resizebox{\textwidth}{!}{
\fontsize{9}{11}\selectfont
\begin{tabular}{lcccccc}
\toprule
    & \textbf{Weapon} & \textbf{Smile} & \textbf{Teeth} & \textbf{Covered face} & \textbf{Dressed Black} & \textbf{Dressed White} \\
\midrule
GPT-4o offer        & 1.3459*** & -1.5389*** & 2.7384*** & 0.6912*** & 2.0340*** & -0.6360*** \\
Haiku offer         & 2.2577*** & -0.1011    & 5.7983*** & 0.4494*** & 2.9914*** & -1.4895*** \\
GPT-4o acceptance   & 0.7262*** & 1.0128***  & 1.3006*** & -0.1757   & 1.4006*** & -0.1896    \\
Haiku acceptance    & 0.6030*** & 1.8215***  & 0.5275**  & -0.6027** & 0.6918*** & -1.4559*   \\
\bottomrule
\end{tabular}
}
\caption{Multiple regression analysis results between objective appearance factors and negotiation outcomes (significant at *p < 0.05; **p < 0.01; ***p < 0.001).}
\label{tab3}
\end{table*}

\section{Experiment setup}
We assigned each avatar image in our dataset to LLMs as a visual persona and asked to participate in a negotiation game. We adopted the negotiation game not to assess the rationality or negotiation skills of LLMs from a game theory perspective, but as a proxy to measure behavior based on the assigned personas. Two experiments were designed to answer the following questions: (1) Can LLMs adjust their behavior based on visual persona? (2) Can LLMs adjust their behavior based on the relative differences between their own and others’ visual traits?

We employed state-of-the-art multi-modal LLMs, specifically OpenAI’s GPT-4o \cite{openai2024gpt4} and Anthropic’s Claude 3 Haiku, for cost-effectiveness. Given the stochastic nature of LLMs, all experiments were conducted at a temperature of 1.0 and repeated five times. All prompts used in the games are documented in \autoref{appendixb}.

\subsection{Study 1: Negotiation Behavior of LLMs Based on Visual Traits}
Following \citet{yee2007proteus}, this study investigates whether LLMs' negotiation behaviors change based on the aggressiveness of the assigned avatar images. LLM is required to play an ultimatum game, where two individuals alternately decide how a pool of money should be split. One participant proposes the split, and the other can either accept or reject it. If accepted, the money is shared as proposed; if rejected, neither participant receives any money. We hypothesized that LLMs with more aggressive images would exhibit more aggressive negotiation behaviors and be more inclined to propose unfair splits compared to those with less aggressive images \cite{yee2007proteus}.

To ensure the consistency of the LLM’s behaviors, the ultimatum game was structured into four rounds. For each image, the assigned LLM participated in two scenarios, acting as either the proposer or the responder in the first round. The total number of negotiation scenarios was 10,370. The LLM was prompted to adopt avatar images representing themselves and negotiate with a confederate. For a fine-grained analysis of the behavioral change depending on the image, the negotiation behavior of the confederate was controlled. The confederate accepted all proposals from the LLM, consistently proposing an initial 50:50 split (in round 1 and 2) and a 75:25 split (in round 3 and 4) in their own favor, allowing for an analysis of the LLM's responses to unfair offers. 

\subsection{Study 2: Negotiation Behavior of LLMs Based on Relative Visual Traits}
This experiment investigated whether LLMs adjust their behavior when simultaneously prompted by images of both themselves and their opponents. In this experiment, two LLMs, each prompted with images representing themselves and their opponents, participate in an ultimatum game. The game is structured into four rounds, with each LLM alternating between proposer and responder (see \autoref{fig1}). Five representative images per aggressiveness score level were used. Subject images were selected based on the results of Study 1, specifically those closest to the average offer amount for each aggression group. The total number of negotiation scenarios was 1,225.

\section{Results}
\subsection{Experiment Results of Study 1}
We investigated how the negotiation behaviors of the LLMs vary according to the aggressiveness of the assigned image. A significant difference was observed between the negotiation results with and without the assignment of a visual persona through images (GPT-4o: t = 20.031, p < 0.001; Claude 3 Haiku: t = 34.309, p < 0.001), indicating that assigning a visual persona through images had a notable  impact on negotiation outcomes. We analyzed the relationship between the LLMs' perception of aggressiveness in each assigned image and the averaged offer amounts across all rounds (GPT-4o: Avg = \$ 63.47, SD = 2.84; Claude 3 Haiku: Avg = \$ 63.13, SD = 4.87) through a linear regression analysis. The results indicate that both GPT-4o and Claude 3 Haiku proposed higher amounts in their favor as the aggressiveness of the assigned image increased (GPT-4o: $\beta$ = 0.8614, p < 0.001, R² = 0.442; Claude 3 Haiku: $\beta$ = 1.657, p < 0.001, R² = 0.287). These findings are consistent with results from previous studies on human-subjects and suggest that LLMs comprehend appearance-based aggressiveness and exhibit aggressive subsequent behaviors \cite{yee2007proteus}.

Next, we analyzed each LLM’s response to unfair offers from the confederate (i.e., only 25\% share for LLMs) using logistic regression analysis. Acceptance probabilities were calculated across all iterations and two negotiation scenarios, coding them as 1 (acceptance) if the probability exceeded 0.5, and 0 (rejection) if the probability was below 0.5. Analysis revealed that both models' acceptance of unfair offers increased with the aggressiveness of their assigned image (GPT-4o: $\beta$ = 0.3643, z = 10.814, p < 0.001, Pseudo R² = 0.06601; Claude 3 Haiku: $\beta$ = 0.1108, z = 2.742, p < 0.001, Pseudo R² = 0.00345). Contrary to the results of human-subject research \cite{yee2007proteus}, LLMs that were assigned more aggressive visual persona were more likely to accept an unfair offer. One possible interpretation is that LLMs with more aggressive persona, having proposed unfair amounts to opponents, tend to be more accepting of unfair offers from opponents as well \cite{kirchsteiger1994role}.

In addition, since LLMs’ behavior is most evident in the generated text \cite{chawla2023selfish,jiang2023personallm}, we conducted a qualitative analysis of the negotiation dialogues. Using the emotion analysis tool LIWC-22\footnote{\url{https://www.liwc.app/}}, we analyzed the negotiation dialogue of each model (see \autoref{tab2}). The analysis revealed similar emotional patterns in the text generated by both models. For instance, models with lower-aggression images used the term 'we' more frequently than 'I' or 'you,' \cite{simmons2005pronouns,kern2012bridging} whereas increased aggression in the image persona led to a reduction in positive tone, politeness, and prosocial behavior, as well as an increase in negative tone and conflict in the generated dialogue. These tendencies support our quantitative analysis results, suggesting that each model comprehends its assigned visual persona and generates dialogue aligned with it.

\begin{figure}
    \centering
    \includegraphics[width=1\linewidth]{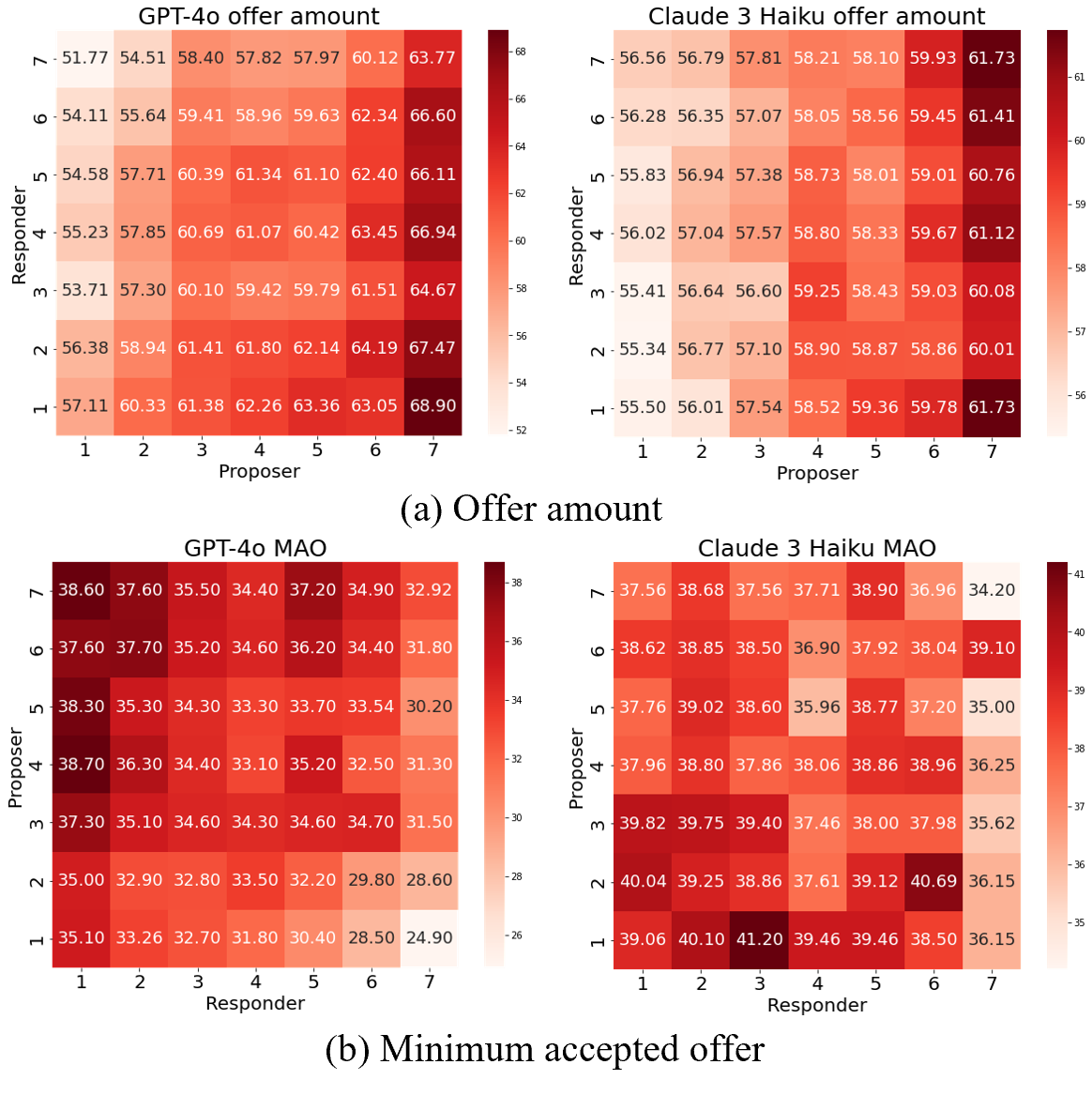}
    \caption{Heatmap of (a) offer amounts and (b) Minimum accepted offer are based on LLMs’ own aggressiveness and the opponent's aggressiveness.}
    \label{fig2}
\end{figure}

We also conducted a fine-grained analysis of the impact of each image’s visual factors on negotiation outcomes (see \autoref{tab3}). The analysis revealed that, similar to the results in \autoref{tab1}, visual factors were correlated with aggression regarding each model’s offer amount. However, for acceptance, the smile factor had the opposite effect compared to the results in \autoref{tab1}. Specifically, while \autoref{tab1} showed that smiles were associated with lower aggression, and lower aggression led to a lower acceptance rate, the analysis indicated that smiles contributed to a higher acceptance rate. This suggests that each model interpreted the smile, not as a signal of lower aggression, but rather as a cue linked to more agreeable behavior (i.e., approving the opponent’s offer).

\subsection{Experiment Results of Study 2}
Next, we examined whether LLMs adjust their negotiation behavior not only based on their own image's aggressiveness but also considering the aggressiveness of their opponent’s image. Multiple regression analysis was conducted using both the average of all-around offer amounts and the perceived aggressiveness of the images assigned to the LLMs and their opponents. The results showed that GPT-4o's offer amounts varied depending on both its own and its opponent's aggressiveness (own aggressiveness: $\beta$ = 1.617, p < 0.001; opponent’s aggressiveness: $\beta$ = -0.614, p < 0.001; R² = 0.473), whereas Claude 3 Haiku’s offer amounts were only influenced by its aggressiveness (own aggressiveness: $\beta$ = 0.789, p < 0.001; opponent’s aggressiveness: $\beta$ = 0.030, p = 0.341; R² = 0.334). As shown in \autoref{fig2} (a), GPT-4o increased its offers as its aggressiveness increased but decreased as its opponent's aggressiveness increased. For instance, a GPT-4o with an aggressiveness level of 1 offered an average of \$ 57.106 to a same level opponent but reduced the offer to \$ 51.77 against a level 7 opponent. In contrast, GPT-4o with an image at aggressiveness level 7 offered \$ 68.9 against a level 1 opponent, decreasing to \$ 63.77 when facing another level 7 opponent. This indicates that GPT-4o adjusts its behavior based on the relative aggressiveness between its image and that of its opponent. This result suggests that GPT-4o may behave submissively toward stronger opponents and more aggressively toward weaker ones, much like humans do \cite{festinger1954theory,debove2016models}. However, the Claude 3 Haiku did not appear to adequately consider the aggressiveness of the opponent, basing negotiations solely on its aggressiveness. Interestingly, this pattern changes as the negotiation progresses. In the case of Claude 3 Haiku, if its previous offer was rejected, its next proposal is influenced not only by its own aggressiveness but also by the aggressiveness of the opponent (own aggressiveness: $\beta$ = 0.6230, p < 0.001; opponent’s aggressiveness: $\beta$ = -0.1679, p < 0.01; R² = 0.019). This suggests that after experiencing a rejection, the model shifts from considering only its own traits to also taking into account the traits of the opponent (i.e., becoming more attuned to the counterpart's aggressiveness).

Furthermore, we analyze the LLMs’ responses to their opponents’ offers by evaluating the minimum accepted offer (MAO), which is the lowest offer amount accepted by a participant during negotiations. Unlike Study 1, the opponent’s offers were not controlled; therefore, we focused on analyzing MAO rather than acceptance probability \cite{chang2011learning}. Multiple regression analysis was performed with the LLMs’ and their opponents’ image aggressiveness as independent variables and MAO as the dependent variable. Both GPT-4o and Claude 3 Haiku showed variations in MAO dependent on their own and their opponents' aggressiveness (GPT-4o: own aggressiveness: $\beta$ = 0.741, p < 0.001; opponent’s aggressiveness: $\beta$ = -0.958, p < 0.001; R² = 0.229; Claude 3 Haiku: own aggressiveness: $\beta$ = -0.259, p < 0.001; opponent’s aggressiveness: $\beta$ = -0.345, p < 0.001; R² = 0.028). GPT-4o's MAO increases with its aggressiveness, indicating that it seeks higher offers as shown in \autoref{fig2} (b). However, it tends to accept lower offers as its opponent's aggressiveness increases, showing reluctance to reject offers from more aggressive opponents. Claude 3 Haiku exhibited a negative influence from both its own and its opponent's aggressiveness toward the MAO. This result could be due to Claude 3 Haiku considering only its own aggressiveness when making offers. Similar to the findings from Study 1, Claude 3 Haiku has tended to make increasingly unfair proposals as its own aggressiveness increases, thereby becoming more likely to accept unfair offers (acting to lower the MAO). At the same time, more aggressive opponents are likely to make more unfair offers, which also contributes to lowering the MAO. 

\section{Conclusion}
This study is the first to explore whether Large Language Models (LLMs) can embody and behave according to a visual persona based on appearance characteristics provided through images. We developed a novel dataset of virtual avatar images to assign visual personas to LLMs and found that LLMs interpret these image-based characteristics in a manner similar to humans. Our results revealed that LLMs recognize aggressiveness based on appearance and that these traits influence their behaviors. Notably, GPT-4o not only understands its own appearance traits but also those of its opponents, adjusting its behavior based on these relative differences—mirroring human tendencies to dominate less aggressive counterparts and submit to more aggressive ones.

\section*{Limitations}
\textbf{Model} We employed state-of-the-art multi-modal LLMs, GPT-4o and Claude 3 Haiku, which are among the models requiring the lowest API costs currently available. Despite this, the costs of our experiments amounted to approximately \$2,625 for GPT-4o and \$212 for Claude 3 Haiku. We hope future research will explore other models (e.g., Claude 3 Sonnet, Opus) more extensively.

\textbf{Experiment} We investigated whether LLMs can understand visual personas and generate aligned outputs; however, there are still areas that require further research. First, we analyzed the negotiation behaviors of LLMs assigned with visual personas specifically in terms of aggressiveness. Although the aggressiveness of appearances is one of the most intuitive elements to explain negotiation behavior \cite{johnson1979deindividuation,yee2007proteus}, the influence of other appearance factors also needs to be explored. In particular, additional exploration is necessary to better understand the acceptance of unfair offers by LLMs (section 4.1) and the MAO for Claude 3 Haiku (section 4.2), which differ from results of existing human-subject research. We anticipate that future studies will utilize the diverse visual traits present in our virtual avatar image dataset for a multifaceted exploration. Second, while we implemented negotiations exclusively between the same models (GPT-4o, Claude 3 Haiku), future research should explore negotiations between different models. This would enable more complex analyses based on each model's understanding of its persona and behavioral patterns, leading to a deeper comprehension of LLM behavior.

\section*{Ethical Considerations}
We strictly adhere to the ACL Code of Ethics for human annotator employment, comply with regional legal requirements, and has been approved by the Institutional Review Board (IRB). We follow the terms of use released by OpenAI and Anthropic. While we made efforts to filter out excessively violent and sexual images during data collection, the fantasy-like nature of our images means that offensive and violent content may still be included. Caution is advised when using this dataset.

Our work has demonstrated the potential for LLMs to act based on visual personas, contributing to the development of more interactive and human-aligned AI agents. However, our findings also reveal that LLMs may exhibit more aggressive behaviors toward less aggressive counterparts without any instructions, which could potentially have harmful impacts. Therefore, conditioning AI agents' behaviors through visual personas must be explored further from ethical and safety perspectives, considering the potential for misuse.

\section*{Acknowledgements}
This paper was supported by SKKU Global Research Platform Research Fund, Sungkyunkwan University, 2022–2024 and the BK21 FOUR (Fostering Outstanding Universities for Research) funded by the Ministry of Education (MOE, Korea) and National Research Foundation of Korea (NRF).

\bibliography{custom}
\clearpage
\appendix
\renewcommand{\thefigure}{A\arabic{figure}}
\renewcommand{\thetable}{A\arabic{table}}
\setcounter{figure}{0}
\setcounter{table}{0}

\begin{figure*}[t]
    \centering
    \includegraphics[width=1\linewidth]{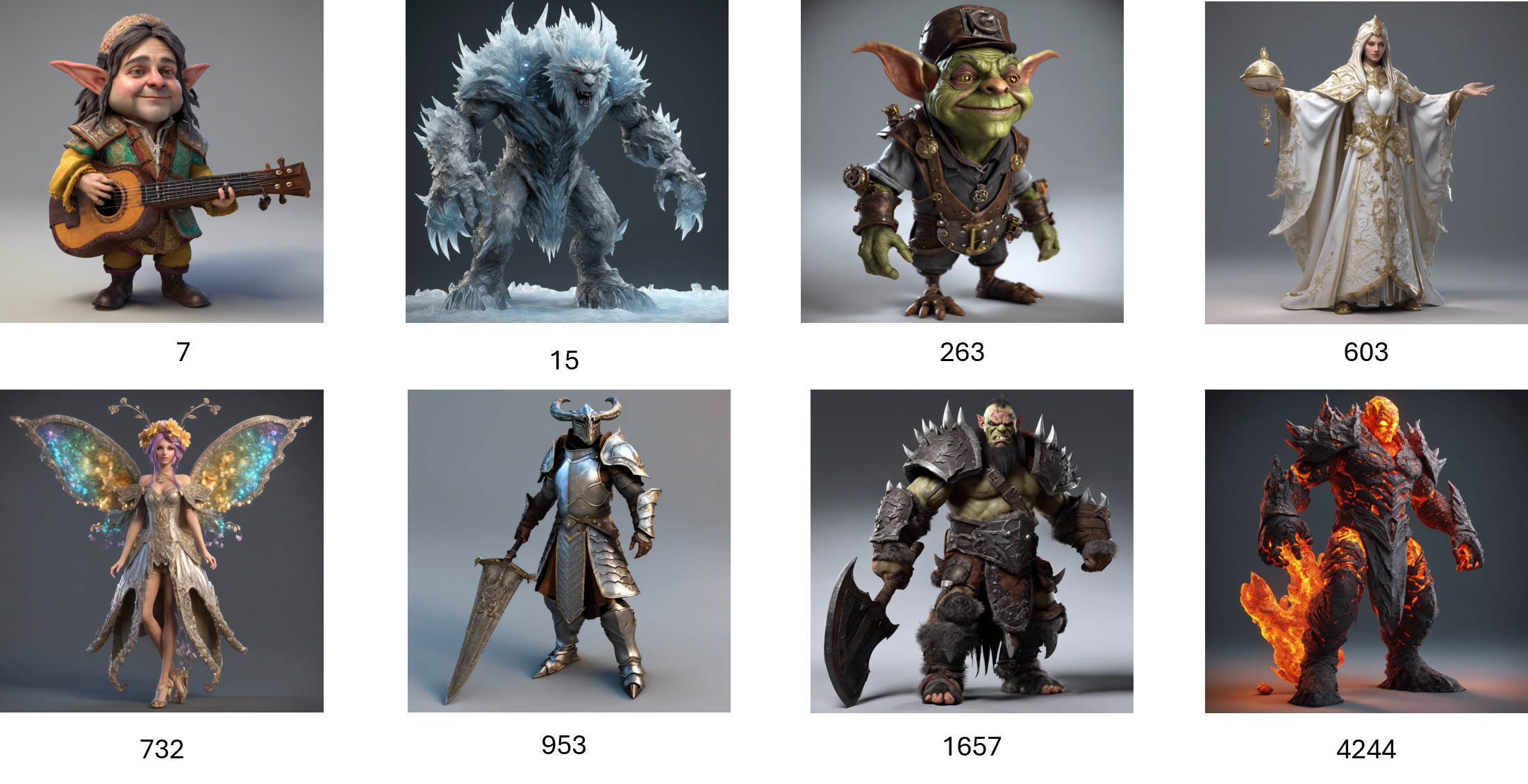}
    \caption{Examples of data}
    \label{figa1}
\end{figure*}

\begin{table*}[t]
    \centering
    \resizebox{1\textwidth}{!}{
    \begin{tabular}{>{\raggedright\arraybackslash}p{3cm}ccccccccc}
         \toprule
         \textbf{Data No.}&  \textbf{Human}&  \textbf{GPT-4o}&  \textbf{Claude 3 Haiku}&  \textbf{Weapon}&  \textbf{Smile}&  \textbf{Teeth}&  \textbf{Covered face}&  \textbf{Dressed Black}& \textbf{Dressed White}\\
         \midrule
         7&  1&  1&  2&  0&  1&  0&  0&  0& 0\\
         15&  6.4&  7&  7&  1&  0&  1&  0&  0& 0\\
         263&  1.4&  1&  3&  0&  1&  0&  0&  0& 0\\
         603&  2&  1&  2&  0&  0&  0&  0&  0& 1\\
         732&  1.4&  1&  2&  0&  1&  0&  0&  0& 0\\
         953&  4.8&  6&  6&  1&  0&  0&  1&  0& 0\\
         1657&  5.4&  7&  7&  1&  0&  1&  0&  0& 0\\
         4244&  6.8&  7&  7&  1&  0&  0&  0&  1& 0\\
         \bottomrule
    \end{tabular}
    }
    \caption{Examples of Data Labeling}
    \label{taba1}
\end{table*}

\begin{table*}[t]
    \centering
    \resizebox{\textwidth}{!}{
    \begin{tabular}{l>{\centering\arraybackslash}p{2cm} >{\centering\arraybackslash}p{2cm} >{\centering\arraybackslash}p{2cm} >{\centering\arraybackslash}p{2cm} >{\centering\arraybackslash}p{2cm} >{\centering\arraybackslash}p{2cm} >{\centering\arraybackslash}p{2cm} >{\centering\arraybackslash}p{2cm} >{\centering\arraybackslash}p{2cm}}
        \toprule
        & \textbf{Human rating} & \textbf{GPT-4o} & \textbf{Haiku} & \textbf{Weapon} & \textbf{Smile} & \textbf{Teeth} & \textbf{Covered face} & \textbf{Dressed Black} & \textbf{Dressed White} \\
        \midrule
        & 3.821 & 3.991 & 5.171 & 0.589 & 0.050 & 0.085 & 0.191 & 0.107 & 0.057 \\
        \bottomrule
    \end{tabular}
    }
    \caption{Dataset Statistics}
    \label{taba2}
\end{table*}

\appendix
\section*{Appendix}
\section{Dataset}
\renewcommand{\thefigure}{\thesection\arabic{figure}}
\renewcommand{\thetable}{\thesection\arabic{table}}
\setcounter{figure}{0}
\setcounter{table}{0}
\renewcommand{\thesubsection}{A.\arabic{subsection}}
\subsection{Dataset Curating}
\label{sec:appendixa1}
To generate virtual avatar images, we utilized the Stable Diffusion model. To craft prompts for the Stable Diffusion model, we employed ChatGPT (GPT-4o). We input the following template into ChatGPT to generate descriptions of fantasy characters randomly: [Randomly create fantasy characters. I wish it was a {race/class} that felt 
{e.g., Threatening, friendly, neutral} with {equipment, armor, weapons, colors, etc.}. Draw it in high detail with a full body image.] The generated prompts were then used in ‘DreamStudio\footnote{\url{https://beta.dreamstudio.ai/generate}},’ a service providing image generation based on the Stable Diffusion model. The images were created with the following settings: Style: ‘3D model,’ Size: 1:1 (1024 × 1024), Negative prompt: \textit{‘Do not draw with the face, torso, or upper body. Never draw background.’}. Ten images were produced for each prompt, and only those featuring a full body were collected for a total of 5,924 images. Subsequently, we filtered 739 images that did not meet our standardized criteria for background, composition, and pose, or that contained sexual or excessive physical errors. Ultimately, 5,185 images were selected to compose the dataset. Examples from this dataset are shown in \autoref{figa1}, \autoref{taba1}, and \autoref{taba2}.

\subsection{Data Annotating}
\label{sec:appendixa2}
To assess how LLMs perceive the aggressive appearance of images, we utilized GPT-4o and Claude 3 Haiku to rate each image's aggressiveness on a scale of 1 to 7. The following prompt was used for the rating process: [Based on the picture of the fictional character, evaluate their appearance. Rate how aggressive the character looks from 1 to 7, with 7 being the most aggressive. Please answer with numbers only.] 

Subsequently, we employed ten graduate students (8 male, 2 female, average age = 27) to annotate the aggressiveness scores of the images. Each annotator was instructed to assess how aggressive the images in the dataset appeared, receiving the same instructions as those provided to the LLMs. On average, annotators took three days to complete the task and were compensated \$50 each. According to their self-reporting, annotators processed an average of 1,350 images per hour.  The annotation results showed a strong inter-annotator correlation, with a minimum correlation coefficient of 0.61 and an average of 0.767, indicating high consistency among the annotators. As described in Section 2, the average scores from the average human ratings and the ratings from each LLM are highly correlated.

\subsection{Objective Appearance Factors}
\label{appendixa3}
To delve deeper into how LLMs perceive the aggressiveness of images, we investigated the impact of objective appearance factors on aggressiveness score rating. A weapon was labeled '1' if it was held in the character’s hand (excluding those slung over the back or sheathed). Hands or forelimbs that appeared sharp and aggressive, such as claws or spikes, were also considered weapons and labeled '1.' Visible teeth were marked '1.' Any form of a smile (closed, upper, and broad) was labeled '1' \cite{otta1996reading} without distinguishing the perception of the stimulus (e.g., smirking versus smiling; both were coded as '1'). Covered faces, whether by a helmet, a mask that fully covered the face, or a hood pushed deep enough to obscure the eyes, were labeled '1.' Black and White were determined based on the clothing worn by the character.
\section{Experiment Details}
\label{appendixb}
\renewcommand{\thefigure}{\thesection\arabic{figure}}
\renewcommand{\thetable}{\thesection\arabic{table}}
\renewcommand{\thesubsection}{B.\arabic{subsection}}
\subsection{Prompts}
\setcounter{figure}{0}
\setcounter{table}{0}

All experiments were repeated five times at a temperature setting of 1.0, and the remaining API settings used each model's default parameters. The values used for result analysis are the averages of these five repetitions. In Study 1, we assigned images to LLMs and had them engage in an ultimatum game with a confederate who followed a fixed script. Each model received its own representative image along with an explanation of the rules of the ultimatum game. The prompt [===round\{round number\}===] was used at the start of each round to help the LLMs differentiate between rounds. All the inputs, including their own images, game rules, round notifications, the confederate's script, and LLMs’ outputs, were cumulatively prompted to the LLMs. Examples of the initial and cumulative prompts provided to the LLMs throughout the experiment are presented in \autoref{tabb2} and \autoref{tabb3}.

\begin{figure}[t]
    \centering
    \includegraphics[width=1\linewidth]{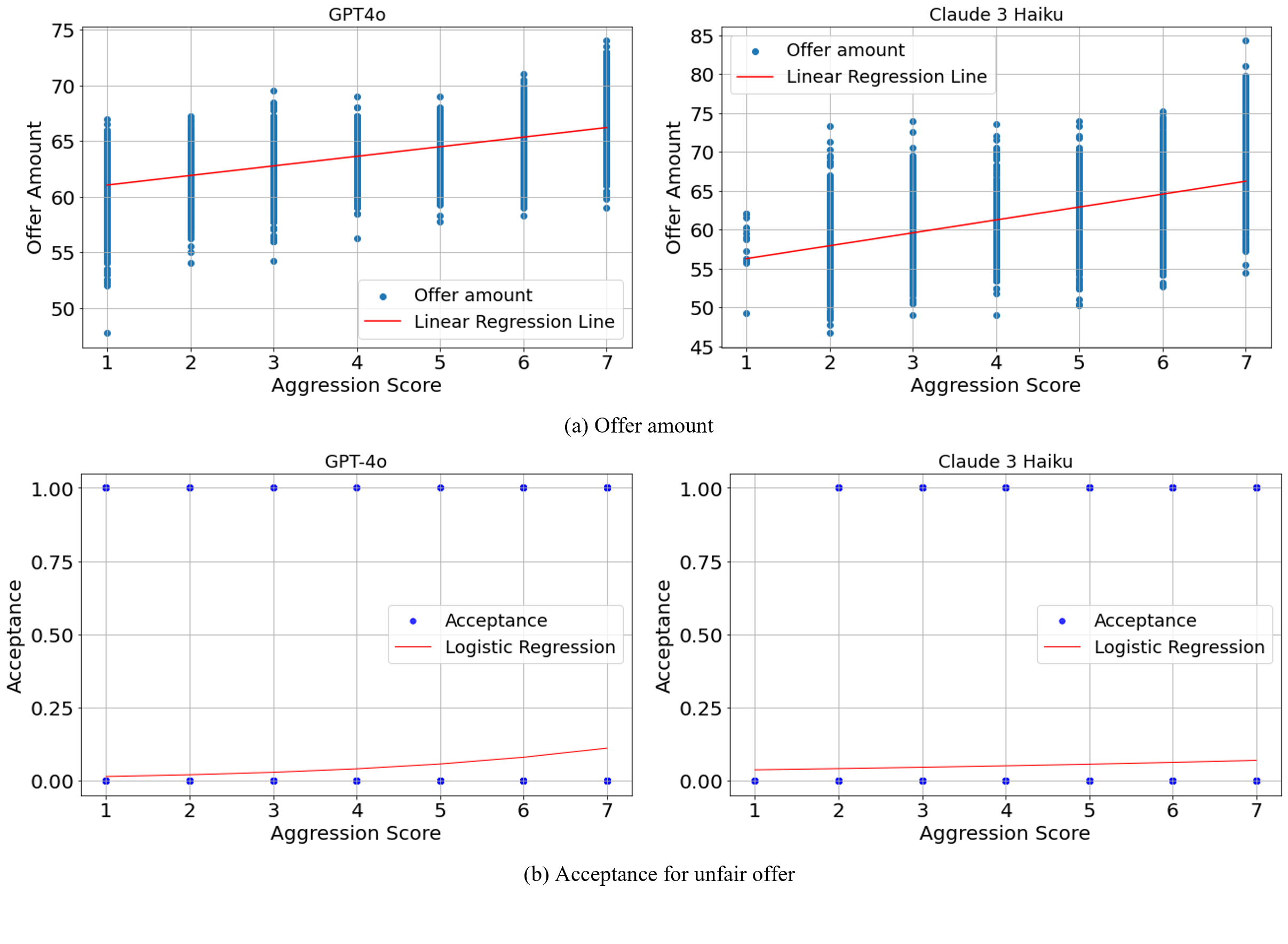}
    \caption{Results of Study 1. Panel (a) shows the results of the regression analysis for offer amounts, and panel (b) displays the results of the logistic regression for the acceptance of unfair offers.}
    \label{figb1}
\end{figure}
\begin{table}[t]
    \centering
    \resizebox{0.5\textwidth}{!}{
    \begin{tabular}{ccc}
        \toprule
        \ \textbf{Agg rating} & \textbf{GPT-4o} & \textbf{Claude 3 Haiku} \\
        \midrule
        1 & 263, 1998, 2662, 3992, 4712 & 851, 2166, 4308, 4839, 4995 \\
        2 & 1607, 3214, 3659, 4178, 5080 & 1410, 2554, 2657, 3905, 4449 \\
        3 & 336, 1616, 2429, 2588, 3228 & 687, 2488, 3275, 4305, 4490 \\
        4 & 709, 2181, 2259, 4089, 4284 & 2483, 3418, 3700, 3763, 3781 \\
        5 & 978, 2913, 4167, 4230, 4282 & 539, 1397, 2351, 3037, 4083 \\
        6 & 448, 839, 3704, 4108, 4514 & 365, 2312, 2786, 4371, 4953 \\
        7 & 93, 483, 1400, 1657, 2363 & 255, 313, 758, 4244, 4480 \\
        \bottomrule
    \end{tabular}
    }
    \caption{Images Used in Study 2}
    \label{tabb1}
\end{table}

In Study 2, both LLMs were given their own and their opponent’s images before engaging in the ultimatum game. As shown in \autoref{tabb1}, five images representing each level of aggressiveness were randomly selected from those closest to the average offer amounts for each aggression interval. Each model received its own representative image, followed immediately by its opponent’s image. Similar to Study 1, their image, the opponent’s image, game rules, round notifications, and all dialogues were cumulatively provided as prompts in each round. Examples of initial and cumulative prompts used during the experiment are listed in \autoref{tabb4} and \autoref{tabb5}. 

\subsection{Analysis of Results}
At the end of each round, the LLMs generated text that outlined their negotiation behavior, and we recorded the data, including the offer amounts and decisions to accept or reject offers. \autoref{figb1} illustrates the visualization of the results analysis from Study 1.
\clearpage
\onecolumn

\begin{figure}
    \centering
    \includegraphics[width=1\linewidth]{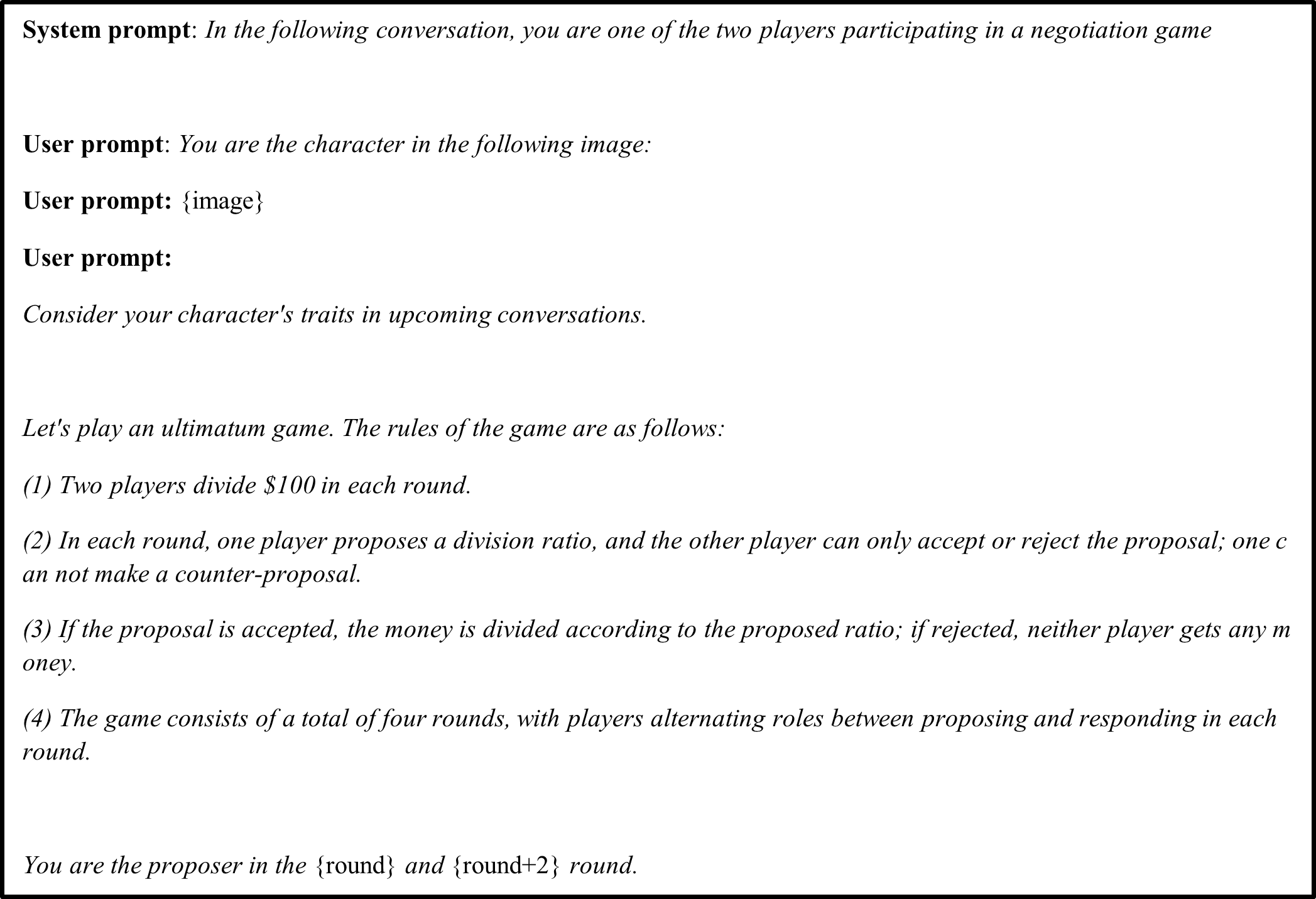}
    \caption{Initial prompts for study 1}
    \label{tabb2}
\end{figure}
\clearpage
\begin{figure}
    \centering
    \includegraphics[width=1\linewidth]{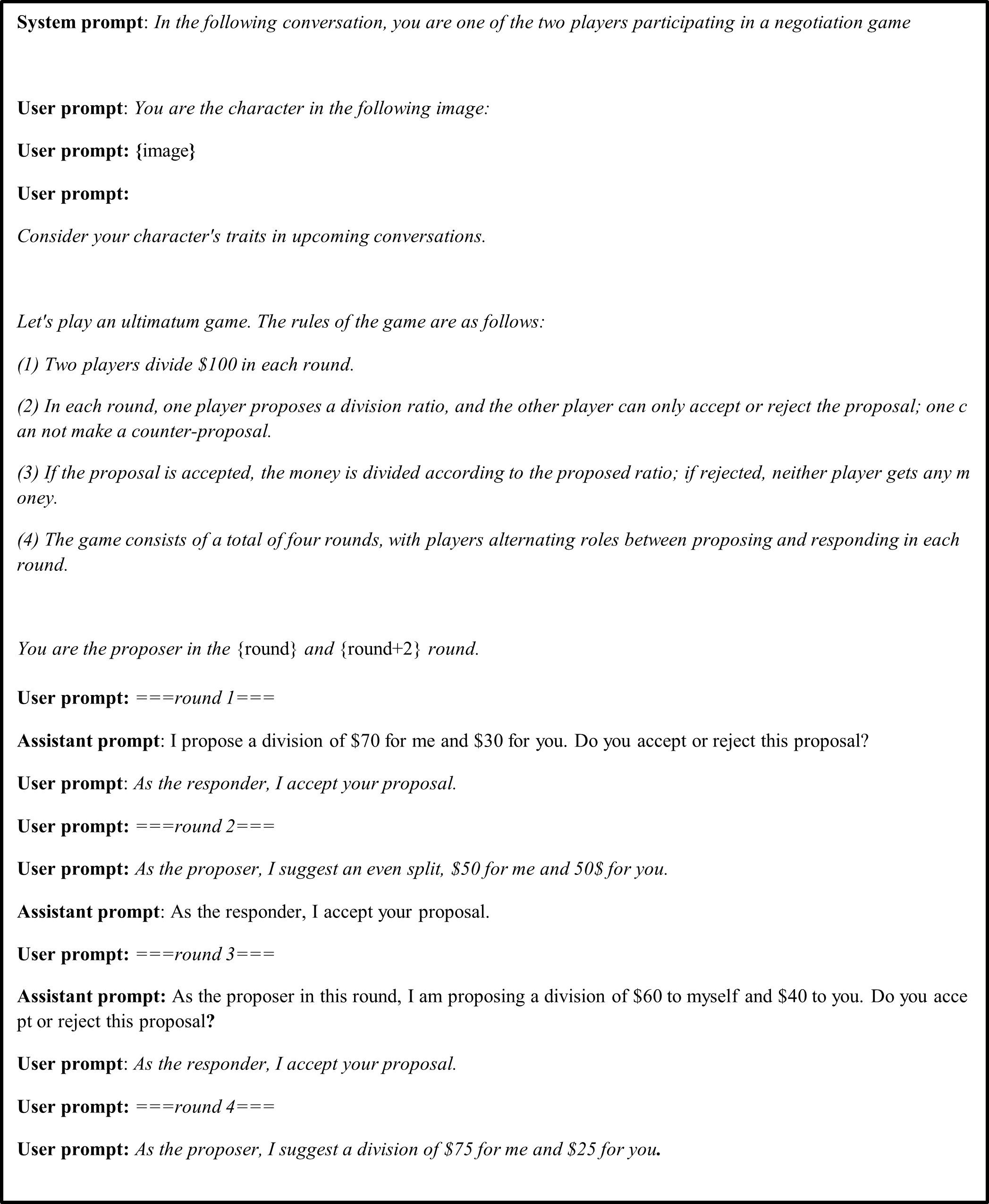}
    \caption{Examples for cumulated prompts for study 1 (GPT-4o) in round 4. Italicized prompts represent those pre-written by researchers, while non-italicized text indicates responses from LLMs.}
    \label{tabb3}
\end{figure}
\clearpage
\begin{figure}
    \centering
    \includegraphics[width=1\linewidth]{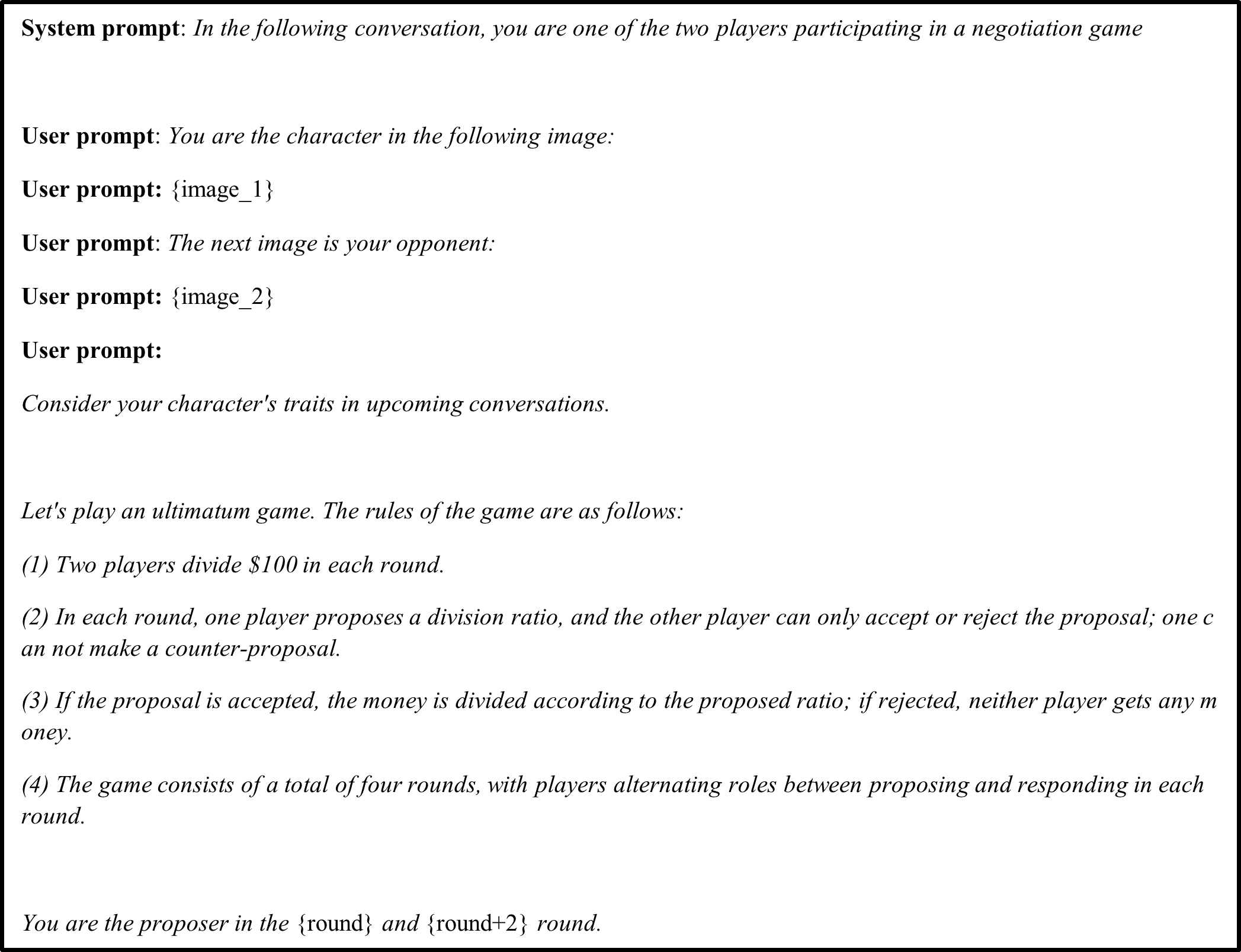}
    \caption{Initial prompts for study 2}
    \label{tabb4}
\end{figure}
\clearpage
\begin{figure}
    \centering
    \includegraphics[width=1\linewidth]{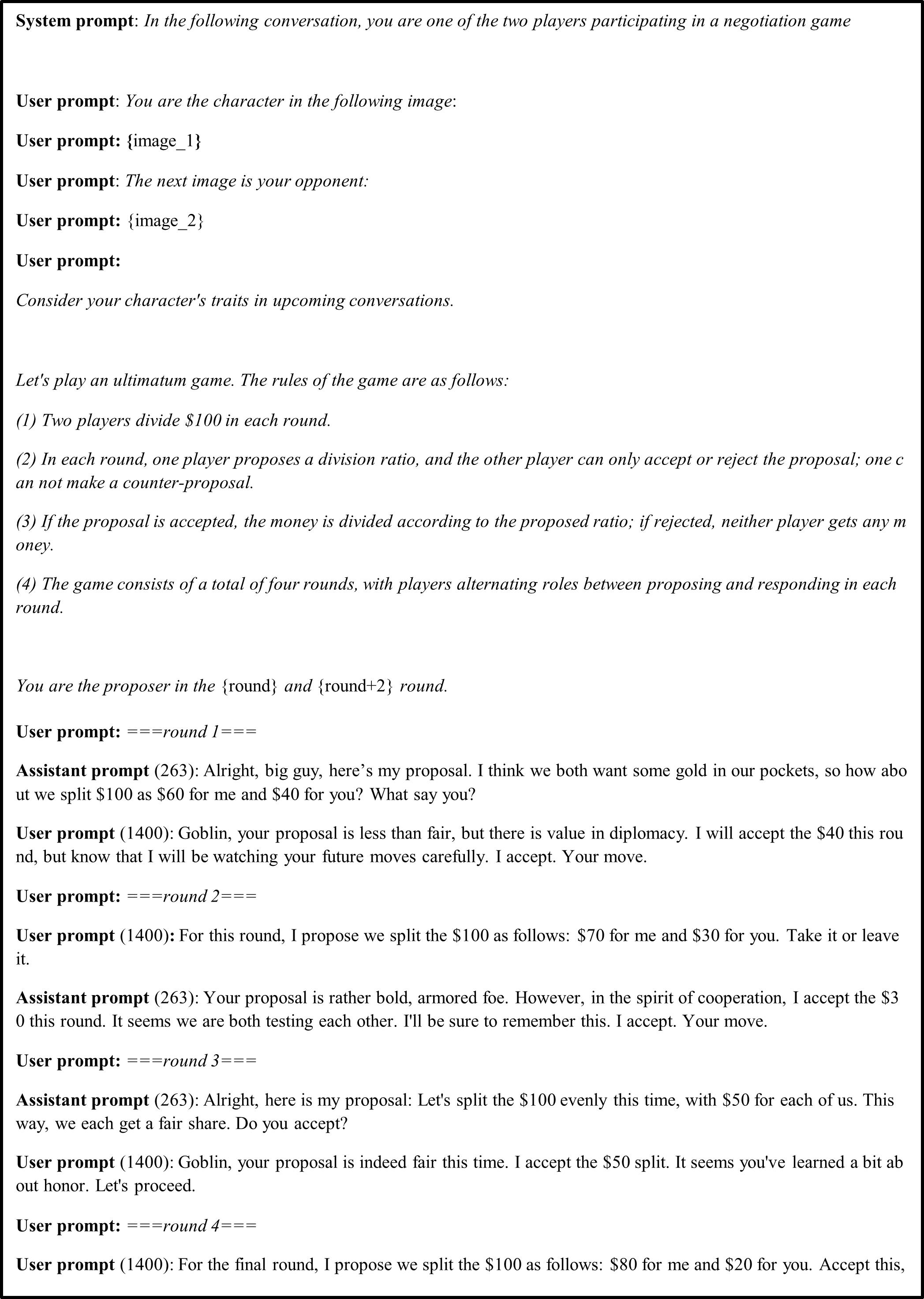}
    \caption{Examples of cumulative prompts for Study 2 (GPT-4o) from the perspective of image 263 (Goblin, Aggressiveness=1) in round 4, versus image 1400 (Black Knight, Aggressiveness=7). Italicized prompts represent those pre-written by researchers, while non-italicized text indicates responses from LLMs.}
    \label{tabb5}
\end{figure}
\end{document}